\documentclass[runningheads]{llncs}

\usepackage[T1]{fontenc}
\usepackage{graphicx}
\usepackage{amssymb}
\usepackage{hyperref}
\usepackage{dsfont}
\usepackage{epstopdf}

\usepackage{floatrow}
\newfloatcommand{capbtabbox}{table}[][\FBwidth]

\usepackage{subcaption} % 用于创建子标题

\usepackage{amsmath}
\usepackage{multirow}
\usepackage{soul}
\usepackage{booktabs}
\usepackage{bbding}
\usepackage{makecell}
\usepackage{stfloats}
\usepackage{xcolor}
\usepackage{makecell}
\usepackage{xspace}
\usepackage{url}
\usepackage{wrapfig}
\usepackage{multirow}
\usepackage{makecell}
\usepackage{booktabs, colortbl}
\usepackage{longtable}
\usepackage{hhline}
\usepackage{algpseudocode}
\usepackage{listings}
\usepackage[ruled]{algorithm2e}
\definecolor{myblue}{HTML}{0072C6}
\definecolor{myyellow}{HTML}{FFFADF}
\definecolor{myred}{HTML}{FF0000}
\usepackage[misc]{ifsym} 

\def\ie{\emph{i.e.}}
\def\eg{\emph{e.g.}}

\def\etc{{\em etc.}} 
\usepackage{rotating}

\usepackage{floatrow}
\begin{document}
% \begin{sloppypar}
%
\title{Diff-VPS: Video Polyp Segmentation via a Multi-task Diffusion Network with Adversarial Temporal Reasoning}
\author{Yingling Lu\inst{1} \and
Yijun Yang\inst{1} \and
Zhaohu Xing \inst{1} \and \\
Qiong Wang\inst{2} \textsuperscript{(\Letter)} \and
Lei Zhu\inst{1,3} \textsuperscript{(\Letter)}
}

% index{Yingling, LU}
% index{Yijun, Yang}
% index{Zhaohu, Xing}
% index{Qiong, Wang}
% index{Lei, Zhu}

\institute{The Hong Kong University of Science and Technology (Guangzhou) \and
{Guangdong Provincial Key Laboratory of Computer Vision and Virtual Reality Technology, Shenzhen Institutes of Advanced Technology, Chinese Academy of Sciences} \and
{The Hong Kong University of Science and Technology} \\
\email{leizhu@ust.hk} 
}
%

% \institute{PaperID 1334}
% \and
% \email{ylu083@connect.hkust-gz.edu.cn}}
%
\maketitle              % typeset the header of the contribution
\begin{abstract}
Diffusion Probabilistic Models have recently attracted significant attention in the community of computer vision due to their outstanding performance. 
However, while a substantial amount of diffusion-based research has focused on generative tasks, no work introduces diffusion models to advance the results of polyp segmentation in videos, which is frequently challenged by polyps' high camouflage and redundant temporal cues.
In this paper, we present a novel diffusion-based network for video polyp segmentation task, dubbed as \textit{Diff-VPS}. 
We incorporate multi-task supervision into diffusion models to promote the discrimination of diffusion models on pixel-by-pixel segmentation. This integrates the contextual high-level information achieved by the joint classification and detection tasks. 
To explore the temporal dependency, Temporal Reasoning Module (TRM) is devised via reasoning and reconstructing the target frame from the previous frames. 
We further equip TRM with a generative adversarial self-supervised strategy to produce more realistic frames and thus capture better dynamic cues. 
Extensive experiments are conducted on SUN-SEG, and the results indicate that our proposed Diff-VPS significantly achieves state-of-the-art performance. 
Code is available at \href{https://github.com/lydia-yllu/Diff-VPS}{https://github.com/lydia-yllu/Diff-VPS}.

\keywords{Diffusion model \and Video polyp segmentation  \and Multi-task learning.}
\end{abstract}

\section{Introduction}

Colorectal cancer (CRC) has become the second leading cause of cancer-related deaths worldwide, accounting for approximately 10\% of all cancer cases. By 2040 the burden of colorectal cancer will increase to 3.2 million new cases per year (an increase of 63\%) and 1.6 million deaths per year (an increase of 73\%)~\cite{bernal2012towards,wu2024colorectal}. Colonoscopy plays a crucial role in detecting polyps and serves as a screening tool for CRC~\cite{xing2024segmamba,xing2022nestedformer}. However, there still exists a high missing rate in the diagnosis and treatment process. 
Recently, emerging methods have utilized deep learning techniques to automatically detect polyps in video, from popular convolutional neural networks to attention mechanism~\cite{tajbakhsh2015automated,puyal2020endoscopic,ji2021progressively,ji2022video,yang2023mammodg,shaharabany2023autosam,ali2024objective,yang2024vivim}. 
Nevertheless, video polyp segmentation (VPS) always poses the ill-posed bottleneck, \ie, the \textit{high camouflage} of polyps in the camera, more specifically, the intrinsic complexity of polyps (shape and color variability, fuzzy boundaries) and the external shooting conditions (low boundary contrast, specular reflection, artifacts). 
Also, the \textit{temporal cues} in colonoscopy videos require to be finely explored, which refers to the effective utilization of long-term temporal characteristics within a video dataset. 
While parts of the above methods achieved significant progress in segmenting polyps in videos, there still exists great potential to be explored considering their failure cases and limited discriminative ability.

Diffusion models have shown impressive performance in many vision tasks, such as image deblurring~\cite{kawar2022denoising}, super-resolution~\cite{li2022srdiff,gao2023implicit}, and anomaly detection~\cite{wyatt2022anoddpm,chen2023diffusiondet}. The denoising diffusion probabilistic model (DDPM)~\cite{ho2020denoising} is a generative model based on a Markov chain. Simulating a diffusion process, DDPM captures the data distribution as it guides the input data towards a predefined target distribution, i.e. Gaussian distribution. 
% Diffusion models can handle image artifacts and can adapt to different image resolutions. 
By representing the diffusion process of image intensity values over successive iterations, diffusion models adeptly grasp the intrinsic structure and texture within images, facilitating the distinct segregation of regions of interest from the background~\cite{amit2023annotator}. 
While several pioneers~\cite{wu2024medsegdiff,xing2023diff,wolleb2022diffusion,yang2023diffmic} applied DDPM to medical image classification and segmentation, there is no diffusion-based work on medical video lesion segmentation.

Motivated by the nature of diffusion models~\cite{ji2023ddp}, in this paper, \textbf{1) we present the first diffusion network, dubbed as Diff-VPS, for the VPS task.} Polyps can be diagnosed into different categories (hyperplastic polyp, invasive carcinoma, \etc) depending on the size and morphological features~\cite{ji2022video}. 
Inspired by this, \textbf{2) we design the Multi-task Diffusion Model where classification and detection tasks are performed simultaneously to improve the discrimination on the segmentation task.} Such supervision significantly advances the discriminative and generalizable ability of diffusion models by high-level semantic information. 
Furthermore, to capture the dynamic appearance and maintain the temporal coherence in segmenting polyps, \textbf{3) we devise a Temporal Reasoning Module by reconstructing the target frame from the previous temporal information and employing an adversarial self-supervised strategy.} 
Extensive experiments are conducted on the SUN-SEG dataset. The superior performance on seen and unseen videos validates the effectiveness and versatility of our Diff-VPS.

%-------------------------------------------------------------------------------------------------------------------
\vspace{-2mm}
\section{Method}

\renewcommand{\floatpagefraction}{.9}
\begin{figure}[!htbp]
\includegraphics[width=\textwidth]{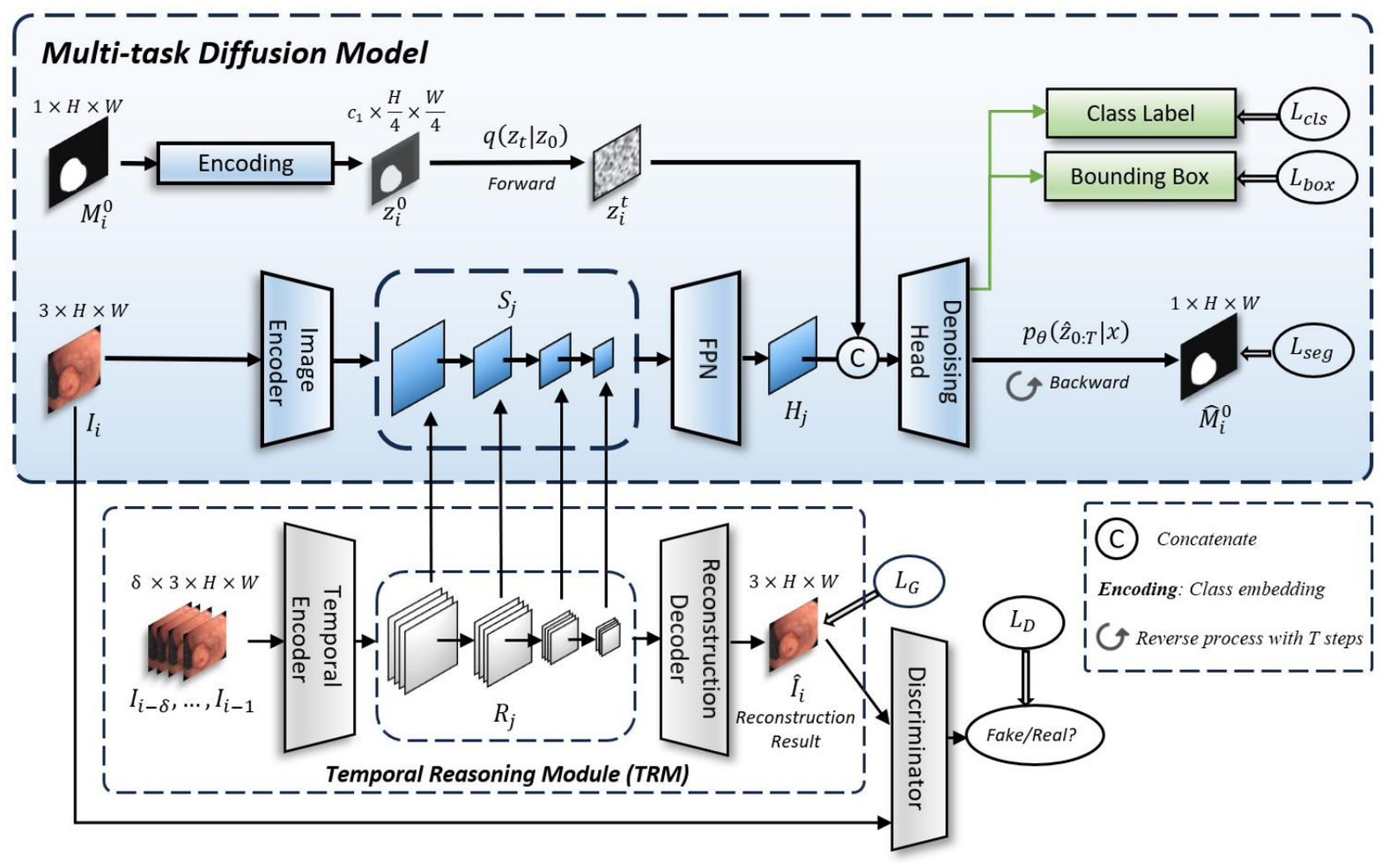}
\caption{\textbf{Overview of our Diff-VPS framework.} While the TRM extracts multi-scale temporal features $\mathcal{R}_j$ from the previous frames, the image encoder learns the spatial counterpart $\mathcal{S}_j$.
The spatiotemporal prior $\mathcal{H}_j$ from $\mathcal{R}_j$ and $\mathcal{S}_j$ conditions the denoising process of our multi-task diffusion model.} 
\label{fig1}
\vspace{-2mm}
\end{figure}

Figure \ref{fig1} shows the schematic illustration of our network for video polyp segmentation. 
Given the target frame $I_i$ and its previous frames $\{I_{i-\delta}, .., I_{i-1}\}$, we first encode the pixel-level mask $M_i^0$ of $I_i$ into the latent space, and then apply diffusion forward process to its latent feature $z_i^0$ by the randomly sampled timestep $t$ to generate the corresponding noisy variable $z_i^t$. 
On the other hand, we feed the target frame $I_i$ into the image encoder to extract its multi-scale spatial features.
Simultaneously, we leverage the previous $\delta$ frames to abstract the multi-scale temporal features by reconstructing the target frame in our Temporal Reasoning Module. 
The temporal features are further integrated with the spatial feature scale-by-scale to construct the spatiotemporal prior. Finally, the denoising head predict the segmentation result $\hat{M}_i^0$ from the noisy variable $z_i^t$ conditioned by the spatiotemporal prior.

\vspace{-2mm}
\subsection{Multi-task Diffusion Model (MDM)} 
\noindent\textbf{Diffusion models. }
Our MDM is developed based on the conditional diffusion model~\cite{ho2020denoising,ji2023ddp}, which belongs to the category of likelihood-based models inspired by non-equilibrium thermodynamics. The conditional diffusion model presupposes a forward noisy process where Gaussian noise is incrementally introduced to the data sample. This process is defined as:

\begin{equation}
q\left(\boldsymbol{z}_t \mid \boldsymbol{z}_0\right)=\mathcal{N}\left(\boldsymbol{z}_t ; \sqrt{\bar{\alpha}_t} \boldsymbol{z}_0,\left(1-\bar{\alpha}_t\right) \mathbf{I}\right) ,
\label{eq:1}
\end{equation}
where $\bar{\alpha}_t:= \prod_{s=0}^t \alpha_s=\prod_{s=0}^t\left(1-\beta_s\right)$ and $\beta_s$ represents the noise scheduler. This process transforms the ground truth variable $z_0$ to a noisy variable $z_t$ by the randomly sampled timestep $t \in[0,T)$, where $z_t \sim \mathcal{N}(0, \mathbf{I})$. 
When training, the reverse process denoising head $f_\theta\left(z_t, \boldsymbol{x}, t\right)$ is trained to predict $z_0$ from $z_t$ under the guidance of condition $\boldsymbol{x}$ by minimizing the objective function. 
When testing, the prediction $\hat{z}_0$ is reconstructed from the Gaussian noise $z_T$ with the model $f_\theta$ conditioned by $\boldsymbol{x}$, in a markovian way $(\boldsymbol{z}_T  \rightarrow \boldsymbol{\hat{z}}_{T-1} \rightarrow \cdots \rightarrow \boldsymbol{\hat{z}}_{T-\Delta} \rightarrow \cdots \rightarrow \hat{z}_0, 0 < \Delta < T)$, which can be formulated as:
\vspace{-2mm}
\begin{equation}
p_\theta\left(\boldsymbol{\hat{z}}_{0: T-1} \mid \boldsymbol{x}\right)=p\left(\boldsymbol{z}_T\right) p_\theta\left(\boldsymbol{\hat{z}}_{T-1} \mid \boldsymbol{z}_T, \boldsymbol{x}\right) \prod_{t=1}^{T-1} p_\theta\left(\boldsymbol{\hat{z}}_{t-1} \mid \boldsymbol{\hat{z}}_t, \boldsymbol{x}\right).
\end{equation}

\noindent\textbf{Multi-task supervision. }
As shown in Fig.~\ref{fig1}, in addition to the main task segmentation, we introduce polyp classification and detection as auxiliary tasks into the conditional diffusion model. 
Classification and detection tasks effectively exploit high-level semantic information from objects in frames, thus providing contextual and discriminative information for segmentation tasks. 
Specifically, the transformer-based image encoder receives the raw target frame $I_i$ as input to generate multi-scale spatial features $\mathcal{S}_j$, specifically with the size of $c_j \times \frac{H}{2^{j+1}} \times \frac{W}{2^{j+1}}, j\in \{1,2,3,4\}$. 
Meanwhile, the transformer-based temporal encoder takes the previous frames $\{I_{i-\delta}, .., I_{i-1}\}$ as input to produce multi-scale temporal features $\mathcal{R}_j$.
Subsequently, $\mathcal{S}_j$ and $\mathcal{R}_j$ are integrated and fed into a feature pyramid network to construct the latent spatiotemporal priors $\mathcal{H}_j$.
To predict the segmentation mask from the noisy mask $z^t_i$, we introduce a CNN-based denoising head $f_\theta$.
The multi-level spatiotemporal features $\mathcal{H}_j$ are concatenated with the noisy variable $z^t_i$ in the latent space to provide conditional guidance for denoising.
Then they are passed into an MLP layer to unify the channel dimension. These unified features are up-sampled to the same resolution and added together.
After that, an MLP layer is adopted to fuse these features.
Finally, the fused feature goes through a $1\times1$ convolutional layer to predict the segmentation mask $\hat{M}_i^0$. 
The denoising head also outputs the classification score $\hat{Y}_{cls}$ and the bounding box $\hat{Y}_{box}$ for instance detection in the target frame $I_i$. 
The segmentation loss $\mathcal{L}_{seg}$ consisting of pixel-wise cross-entropy loss, mean square error and IoU loss is applied during training.
And the two auxiliary tasks are optimized by cross-entropy loss. Thus, the overall objective function of MDM is:
\vspace{-2mm}
\begin{equation}
\mathcal{L}_{MDM}= \lambda_{seg} \cdot \mathcal{L}_{seg} + \lambda_{cls} \cdot \mathcal{L}_{ce} (\hat{Y}_{cls}, Y_{cls}) + \lambda_{det} \cdot \mathcal{L}_{ce} (\hat{Y}_{box}, Y_{box}) ,
\end{equation}
where $\lambda_{seg}, \lambda_{cls}, \lambda_{det}$ are the balancing hyper-parameters, empirically set as 0.5, 0.05, 0.2, respectively. $Y_{cls}$ is the category of polyps in $I_i$, \ie, Low-grade adenoma, High-grade adenoma, Hyperplastic polyp, Traditional serrated adenoma, Sessile serrated lesion, Invasive carcinoma~\cite{ji2022video}; 
$Y_{box}$ is $(x_c, y_c, w, h)$ of a bounding box for the target lesion.
The multi-task supervision strategy critically enhances the discriminative and generalizable ability of conditional diffusion models.
More importantly, object-level annotations mitigate the class-imbalanced teaser in pixel-level segmentation of polyps due to the scarcity of lesions.

%-------------------------------------------------------------------------------------------------------------------
\vspace{-2mm}
\subsection{Temporal Reasoning Module (TRM)} 
To capture the dynamic appearance and motion cues in colonoscopy videos, we design a temporal reasoning module as displayed in Fig.~\ref{fig1}. 
To be specific, we formulate a self-supervised proxy task that reconstructs the target frame $I_i$ from the previous frames $\{I_{i-\delta}, \cdots, I_{i-1}\}$.
We feed the previous frames into the temporal encoder to extract their multi-scale features frame-by-frame.
Then, they are averaged along the temporal axis and aggregated into multi-scale temporal features $\{\mathcal{R}_j \in \mathbb{R}^{\mathrm{c}_j \times \frac{H}{2^{j+1}} \times \frac{W}{2^{j+1}}} \big| j \in [1,4]\}$.
The CNN-based reconstruction decoder consisting of MLP layers receives and integrates multi-scale temporal features to predict the next frame $\hat{I}_i$.
By reasoning the unseen future, the temporal encoder explores the temporal dependency from video sequences and learns the spatiotemporal discriminative features to provide guidance for segmenting polyps with high camouflage.

To produce more realistic frames, we further introduce an adversarial self-supervision on the reconstructed results inspired by \cite{xu2019spatiotemporal}. 
Our designed TRM is treated as the generator $\mathcal{G}$ while the Inception3 network pre-trained on the ImageNet is adopted as the discriminator $\mathcal{D}$ to distinguish the generated frame $\hat{I}_i$ from the real frame $I_i$. 
For the fixed parameter of $\mathcal{G}$, we minimize the adversarial loss $\mathcal{L}_{\mathcal{D}}$ to optimize the discriminator $\mathcal{D}$ as follows:
\begin{equation}
\mathcal{L}_{\mathcal{D}}=-\log(1-D(\hat{I}_i))-\log D(I_i).
\end{equation}
Simultaneously, for the fixed parameters of $\mathcal{D}$, we expect the generator $\mathcal{G}$ can spoof $\mathcal{D}$ to generate the frame more like a real counterpart:
\begin{equation}
\mathcal{L}_{\mathcal{G}}=\mathcal{L}_{MSE}(\hat{I}_i, I_i) - \lambda_{adv} \cdot \log D(\hat{I}_i),
\end{equation}
where $\mathcal{L}_{TRM}=\mathcal{L}_{\mathcal{G}}$ and $\lambda_{adv}$ is set to 0.001 to balance the reconstruction term of mean square error and the adversarial term. 
% Note that $\mathcal{L}_{\mathcal{D}}$ is only adopted to optimize the discriminator, with no respect to TRM.

\noindent\textbf{Total Loss.} By combing the MDM loss and the TRM loss, we compute the total loss $\mathcal{L}_{total}$ of Diff-VPS as follows:

\begin{equation}
\mathcal{L}_{total} = \lambda_{MDM} \cdot \mathcal{L}_{MDM} + \lambda_{TRM} \cdot \mathcal{L}_{TRM} ,
\end{equation}
where the hyper-parameters are empirically set as $\lambda_{MDM}=0.75, \lambda_{TRM}=0.25$.  
More details of the training and inference are displayed in the \textit{Supplementary}.

%-------------------------------------------------------------------------------------------------------------------
\section{Experiment}

\subsection{Datasets and Experimental Setting}
We evaluate our proposed Diff-VPS on the largest video polyp dataset SUN-SEG~\cite{ji2022video}. 
SUN-SEG has 113 colonoscopy videos, including 100 positive cases with 49,136 polyp frames, and 13 negative cases with 109,554 non-polyp frames. 
Following \cite{ji2022video}, we use the re-organized positive cases in SUN-SEG where 14,176 frames are used for training and 24,736 frames are used for testing.
We conduct extensive experiments on two testing sub-datasets (i.e. Easy and Hard) based on data distribution and each sub-dataset is divided into seen and unseen parts, where seen represents the visible case that divides one case into two parts for training and testing (e.g. case 7\_2 for training and case 7\_1 for testing). 

\noindent\textbf{Implementation details. }
The proposed framework was implemented on the Pytorch platform and trained by two NVIDIA 3090 GPUs. 
We trained our model for 15 epochs with a batch size of 16.
% We set the batch size to 24, which takes 4 hours for 15 epochs and reaches convergence after around 10 epochs. 
A video clip of 5 frames with a patch size of 224$\times$224 was fed into the network. 
We adopted the Adam optimizer with the initial learning rate of $1\times10^{-4}$, which was decayed by a polynomial scheduler. 

\begin{table}[!t]
\centering
\caption{Quantitative comparison of two testing datasets with seen colonoscopy scenarios. The best values are highlighted in bold. $\uparrow$ denotes the higher, the better.}
% \vspace{-2mm}
\label{tab1}
\resizebox{0.75\linewidth}{!}{
\begin{tabular}{c|c||cccc|cccc}
\hline
\multirow{2}{*}{Model} &\multirow{2}{*}{Publish} & \multicolumn{4}{c|}{SUN-SEG-Easy} &\multicolumn{4}{c}{SUN-SEG-Hard} \\
 &  & $S_{\alpha}\uparrow$ & $E^{mn}_\phi\uparrow$ & $F^{w}_\beta\uparrow$ & Dice$\uparrow$ & $S_{\alpha}\uparrow$ & $E^{mn}_\phi\uparrow$ & $F^{w}_\beta\uparrow$ & Dice$\uparrow$ \\
\hline
COSNet~\cite{lu2019see} & $TPAMI_{19}$ & 0.845 & 0.836 & 0.727 & 0.804 & 0.785 & 0.772 & 0.626 & 0.725 \\
MAT~\cite{zhou2020matnet} & $TIP_{20}$  & 0.879 & 0.861 & 0.731 & 0.833 & 0.840 & 0.821 & 0.652 & 0.776 \\
PCSA~\cite{gu2020pyramid} & $AAAI_{20}$  & 0.852 & 0.835 & 0.681 & 0.779 & 0.772 & 0.759 & 0.566 & 0.679 \\
2/3D~\cite{puyal2020endoscopic} & $MICCAI_{20}$  & 0.895 & 0.909 & 0.819 & 0.856 & 0.849 & 0.868 & 0.753 & 0.809 \\
AMD~\cite{liu2021emergence} & $NeurIPS_{21}$  & 0.471 & 0.526 & 0.114 & 0.245 & 0.480 & 0.536 & 0.115 & 0.231 \\
DCF~\cite{zhang2021dynamic} & $ICCV_{21}$  & 0.572 & 0.591 & 0.357 & 0.398 & 0.603 & 0.602 & 0.385 & 0.443 \\
FSNet~\cite{ji2021full} & $ICCV_{21}$  & 0.890 & 0.895 & 0.818 & 0.873 & 0.848 & 0.859 & 0.755 & 0.828 \\
PNSNet~\cite{ji2021progressively} & $MICCAI_{21}$  & 0.906 & 0.910 & 0.836 & 0.861 & 0.870 & 0.892 & 0.787 & 0.823 \\
PNS+~\cite{ji2022video} & $MIR_{22}$ & 0.917 & 0.924 & 0.848 & 0.888 & 0.887 & 0.929 & 0.806 & 0.855 \\
\textbf{Diff\mbox{-}VPS} & $OURS_{24}$ & $\mathbf{0.930}$ & $\mathbf{0.966}$ & $\mathbf{0.894}$ & $\mathbf{0.908}$ & $\mathbf{0.900}$ & $\mathbf{0.947}$ & $\mathbf{0.851}$ & $\mathbf{0.868}$ \\
\hline
\end{tabular}}
\vspace{-2mm}
\end{table}

\begin{table}[!t]
\centering
\caption{Quantitative comparison of two testing datasets with unseen colonoscopy scenarios. The best values are highlighted in bold. \dag denotes image segmentation method.}
\vspace{-2mm}
\label{tab2}
\resizebox{0.75\linewidth}{!}{
\begin{tabular}{c|c||cccc|cccc}
\hline
\multirow{2}{*}{Model} &\multirow{2}{*}{Publish} & \multicolumn{4}{c|}{SUN-SEG-Easy} &\multicolumn{4}{c}{SUN-SEG-Hard} \\
 & & $S_{\alpha}\uparrow$ & $E^{mn}_\phi\uparrow$ & $F^{w}_\beta\uparrow$ & Dice$\uparrow$ & $S_{\alpha}\uparrow$ & $E^{mn}_\phi\uparrow$ & $F^{w}_\beta\uparrow$ & Dice$\uparrow$ \\
\hline
COSNet~\cite{lu2019see} & $TPAMI_{19}$ & 0.654 & 0.600 & 0.431 & 0.596 & 0.670 & 0.627 & 0.443 & 0.606 \\
PCSA~\cite{gu2020pyramid} & $AAAI_{20}$ & 0.680 & 0.660 & 0.451 & 0.592 & 0.682 & 0.660 & 0.442 & 0.584 \\
MAT~\cite{zhou2020matnet} & $TIP_{20}$ & 0.770 & 0.737 & 0.575 & 0.710 & 0.785 & 0.755 & 0.578 & 0.712 \\
2/3D~\cite{puyal2020endoscopic} & $MICCAI_{20}$ & 0.786 & 0.777 & 0.652 & 0.722 & 0.786 & 0.775 & 0.634 & 0.706 \\
ACSNet\dag~\cite{zhang2020adaptive} & $MICCAI_{20}$ & 0.782 & 0.779 & 0.642 & 0.782 & 0.783 & 0.787 & 0.636 & 0.708 \\
SANet\dag~\cite{wei2021shallow} & $MICCAI_{21}$ & 0.720 & 0.745 & 0.566 & 0.649 & 0.706 & 0.743 & 0.526 & 0.598 \\
% AMD~\cite{liu2021emergence} & $NeurIPS_{21}$ & 0.474 & 0.533 & 0.133 & 0.266 & 0.472 & 0.527 & 0.128 & 0.252 \\
% DCF~\cite{zhang2021dynamic} & $ICCV_{21}$ & 0.523 & 0.514 & 0.270 & 0.325 & 0.514 & 0.522 & 0.263 & 0.317 \\
PNSNet~\cite{ji2021progressively} & $MICCAI_{21}$ & 0.767 & 0.744 & 0.616 & 0.676 & 0.767 & 0.755 & 0.609 & 0.675 \\
FSNet~\cite{ji2021full} & $ICCV_{21}$ & 0.725 & 0.695 & 0.551 & 0.702 & 0.724 & 0.694 & 0.541 & 0.699 \\
UACANet\dag~\cite{kim2021uacanet} & $ACM MM_{21}$ & 0.831 & 0.856 & 0.754 & 0.757 & 0.824 & 0.848 & 0.734 & 0.739 \\
PNS+~\cite{ji2022video} & $MIR_{22}$ & 0.806 & 0.798 & 0.676 & 0.756 & 0.797 & 0.793 & 0.653 & 0.737 \\
AutoSAM~\cite{shaharabany2023autosam} & $arXiv_{23}$ & 0.815 & 0.855 & 0.716 & 0.753 & 0.822 & 0.866 & 0.714 & 0.759 \\
% IN\cite{dahan2024video} & $MIDL_{23}$ & 0.82 & 0.85 & 0.71 & 0.77 & 0.78 & 0.87 & 0.73 & $\mathbf{0.79}$ \\
\textbf{Diff\mbox{-}VPS} & $OURS_{24}$ & $\mathbf{0.828}$ & $\mathbf{0.883}$ & $\mathbf{0.748}$ & $\mathbf{0.767}$ & $\mathbf{0.823}$ & $\mathbf{0.886}$ & $\mathbf{0.733}$ & $\mathbf{0.764}$ \\
\hline
\end{tabular}}
\vspace{-4mm}
\end{table}

\begin{figure}[!t]
\includegraphics[width=\textwidth]{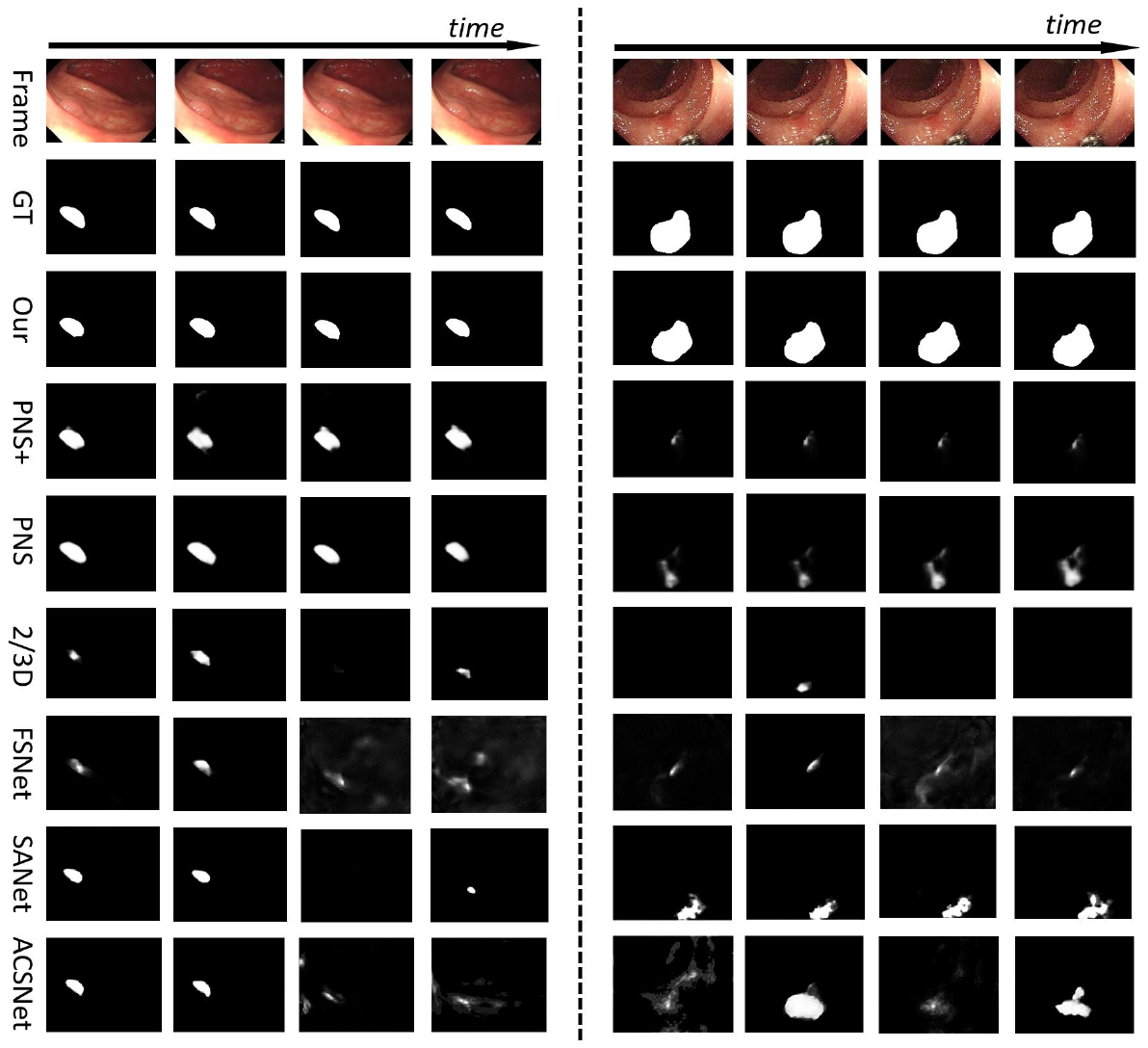}
% \vspace{-2mm}
\caption{Qualitative results on SUN-SEG. Left: a clip from easy-seen dataset case 75. Right: a clip from hard-unseen dataset case 36. 
% From left to right are consecutive frames of a video clip over time. From top to bottom are the frame, annotated mask, and segmentation results from Diff-VPS and other VPS methods.
} \label{fig2}
% \vspace{-4mm}
\end{figure}

\subsection{Comparison with State-of-the-Arts}

\noindent\textbf{Baselines.}
We compare our Diff-VPS with popular image- and video-level object/polyp segmentation methods on SUN-SEG-Easy and SUN-SEG-Hard, \ie, COSNet~\cite{lu2019see}, MAT~\cite{zhou2020matnet}, PCSA~\cite{gu2020pyramid}, 2/3D~\cite{puyal2020endoscopic}, AMD~\cite{liu2021emergence}, DCF~\cite{zhang2021dynamic}, FSNet~\cite{ji2021full}, PNSNet~\cite{ji2021progressively}, PNS+~\cite{ji2022video}, ACSNet~\cite{zhang2020adaptive}, SANet~\cite{wei2021shallow}, UACANet~\cite{kim2021uacanet}, and AutoSAM~\cite{shaharabany2023autosam} in Tab.~\ref{tab1} and \ref{tab2}.

\noindent\textbf{Metrics.}
For a comprehensive evaluation, we use four metrics following \cite{ji2022video}: 
(1) S-measure (${S}_\alpha$), which evaluates region and object aware structural similarity;  %\cite{fan2017structure}
(2) Enhanced-alignment measure ($E_\phi^{mn}$), which measures pixel level matching and image-level statistics; %\cite{fan2018enhanced}
(3) Weighted F-measure ($F_\beta^w$), which amends the “Equal-importance flaw” in Dice and $F_\beta$. %\cite{margolin2014evaluate}
(4) Dice coefficient, which measures the similarity between two sets of data.
% (5) Pixel-wise sensitivity (Sen), evaluates the true positive prediction of overall lesion areas.

% \vspace{-4mm}
\noindent\textbf{Quantitative Comparison.}
As shown in Tab.~\ref{tab1}, our algorithm significantly outperforms the second-best method PNS+ on seen colonoscopy videos across four metrics by a remarkable margin (\eg, for SUN-SEG-Hard, 1.3\% in $S_\alpha$, 1.8\% in $E_\phi^{mn}$, 4.6\% in $F_\beta^w$, 1.2\% in Dice). 
Also, in Tab.~\ref{tab2}, our method is superior to the second-best method AutoSAM on unseen colonoscopy videos, especially for SUN-SEG-Easy (1.3\% in $S_\alpha$, 2.8\% in $E_\phi^{mn}$, 3.2\% in $F_\beta^w$, 1.4\% in Dice).
% We believe that it can be used in our DiffVPS to further improve the performance. 

% \vspace{-2mm}
\noindent\textbf{Qualitative Comparison.}
As displayed in Fig.~\ref{fig2}, we showcase several polyp segmentation results of our Diff-VPS and the comparison methods on the SUN-SEG. Our method demonstrates robust performance in accurately locating and segmenting polyps across diverse and challenging scenarios. It has proficiency in managing polyps of varying dimensions, uniform areas, and textures. The versatility of our model allows it to navigate through intricate situations, providing reliable and precise segmentation results for polyps.

\vspace{-2mm}
\subsection{Ablation Study}

\begin{table}[!t]
\centering
\caption{Ablation studies on SUN-SEG. `MDM' is multi-task diffusion model. `TRM' is temporal reasoning module. `ASS' is adversarial self-supervised strategy.}
\vspace{-2mm}
\label{tab3}
\resizebox{0.8\linewidth}{!}{
\begin{tabular}{c|c|c|c|c|c||cccc}
\hline
\hline
\multirow{2}{*}{Models} & \multirow{2}{*}{MDM} & \multirow{2}{*}{TRM} & \multirow{2}{*}{ASS} & \multirow{2}{*}{Sub-dataset} & \multirow{2}{*}{Scenarios} & \multirow{2}{*}{$S_{\alpha}$} & \multirow{2}{*}{$E^{mn}_\phi$} & \multirow{2}{*}{$F^{w}_\beta$} & \multirow{2}{*}{Dice} \\
 & & & & & & & & \\
\hline
\multirow{4}*{\#1} & \multirow{4}*{} & \multirow{4}*{} & \multirow{4}*{} 
~ & \multirow{2}*{Easy} & Seen & 0.910 & 0.948 & 0.840 & 0.882 \\
~ & & & & & Unseen & 0.797 & 0.850 & 0.736 & 0.726 \\
\cline{5-10}
~ & & & & \multirow{2}*{Hard} & Seen & 0.870 & 0.935 & 0.841 & 0.837 \\
~ & & & & & Unseen & 0.802 & 0.859 & 0.721 & 0.717 \\
\hline
\multirow{4}*{\#2} & \multirow{4}*{$\checkmark$} & \multirow{4}*{} & \multirow{4}*{} 
~ & \multirow{2}*{Easy} & Seen & 0.920 & 0.965 & 0.899 & 0.896 \\
~ & & & & & Unseen & 0.808 & 0.863 & 0.748 & 0.740 \\
\cline{5-10}
~ & & & & \multirow{2}*{Hard} & Seen & 0.882 & 0.932 & 0.853 & 0.839 \\
~ & & & & & Unseen & 0.815 & 0.881 & 0.727 & 0.744 \\
\hline
\multirow{4}*{\#3} & \multirow{4}*{} & \multirow{4}*{$\checkmark$} & \multirow{4}*{} 
~ & \multirow{2}*{Easy} & Seen & 0.917 & 0.968 & 0.896 & 0.895 \\
~ & & & & & Unseen & 0.807 & 0.855 & 0.745 & 0.733 \\
\cline{5-10}
~ & & & & \multirow{2}*{Hard} & Seen & 0.872 & 0.932 & 0.832 & 0.833 \\
~ & & & & & Unseen & 0.807 & 0.870 & 0.726 & 0.725 \\
\hline
\multirow{4}*{\#4} & \multirow{4}*{} & \multirow{4}*{$\checkmark$} & \multirow{4}*{$\checkmark$} 
~ & \multirow{2}*{Easy} & Seen & 0.924 & 0.963 & 0.893 & 0.902 \\
~ & & & & & Unseen & 0.817 & 0.878 & 0.757 & 0.761 \\
\cline{5-10}
~ & & & & \multirow{2}*{Hard} & Seen & 0.874 & 0.937 & 0.844 & 0.847 \\
~ & & & & & Unseen & 0.823 & 0.883 & 0.744 & 0.750 \\
\hline
\multirow{4}*{\textbf{Our method}} & \multirow{4}*{$\checkmark$} & \multirow{4}*{$\checkmark$} & \multirow{4}*{$\checkmark$} 
~ & \multirow{2}*{Easy} & Seen & $\mathbf{0.930}$ & $\mathbf{0.966}$ & $\mathbf{0.894}$ & $\mathbf{0.908}$ \\
~ & & & & & Unseen & $\mathbf{0.828}$ & $\mathbf{0.883}$ & $\mathbf{0.748}$ & $\mathbf{0.767}$ \\
\cline{5-10}
~ & & & & \multirow{2}*{Hard} & Seen & $\mathbf{0.900}$ & $\mathbf{0.947}$ & $\mathbf{0.851}$ & $\mathbf{0.868}$ \\
~ & & & & & Unseen & $\mathbf{0.823}$ & $\mathbf{0.886}$ & $\mathbf{0.733}$ & $\mathbf{0.764}$ \\
\hline
\hline
\end{tabular}}
\vspace{-4mm}
\end{table}

Extensive experiments are conducted on SUN-SEG in Tab.~\ref{tab3} to evaluate the effectiveness of our major components. To achieve this, we construct four baseline networks from our method.  
The first baseline (denoted as ``\#1'') is to remove multi-task supervision and temporal reasoning module from our network. It means that ``\#1'' equals the vanilla diffusion model~\cite{ji2023ddp}.
Then, we introduce classification and detection supervisions into ``\#1'' to construct ``\#2'', and build ``\#3'' by incorporating our TRM without adversarial self-supervision. 
Based on ``\#3'', adversarial self-supervision is applied to construct ``\#4''.
Hence, ``\#4'' equals removing the multi-task supervision from our full method.

\noindent\textbf{Effectiveness of MDM.}
Observed from ``\#1'' and ``\#2'', the MDM strategy significantly improves the performance across all metrics. 
It indicates that category imbalance in pixel-level classification (i.e. segmentation) can be mitigated by applying detection and classification tasks. 
The categories of the detection and classification tasks are relatively balanced as they leverage object-level annotations for supervision. 
The multi-task supervision improves the generalizable ability of the model, which is also proved by incorporating MDM into \#4 to construct our full method.

\noindent\textbf{Effectiveness of TRM.}
As shown in Tab.~\ref{tab3}, ``\#3'' advances the performance of ``\#1'' on seen and unseen videos, indicating that TRM plays a pivotal role in exploiting the temporal redundancy in colonoscopy videos. 
Compared to ``\#3'', ``\#4'' achieves a considerable improvement by adapting the model to the real data distribution and highlighting the perceptual quality of the samples. 
This demonstrates the effect of the generative adversarial self-supervision.

\section{Conclusion}
This paper presents the first diffusion-based multi-task framework for video polyp segmentation (VPS). 
The main idea is to improve the discriminative and generalizable ability of diffusion models by incorporating the multi-task supervision strategy. 
We effectively explore the temporal dependency by incorporating temporal reasoning module to mitigate the high camouflage of polyps in videos.
Experiments demonstrate that our Diff-VPS is capable of state-of-the-art results on VPS benchmark dataset SUN-SEG, which can be a critical baseline for diffusion models on video object/lesion segmentation.

\noindent\textbf{Acknowledgments} The work was supported by National Key R\&D Program of China (Grant No.2023YFB4705700), Guangzhou-HKUST(GZ) Joint Funding Program (No. 2023A03J0671), NSFC General Project  62072452, and the Regional Joint Fund of Guangdong under Grant 2021B1515120011. 

\noindent\textbf{Disclosure of Interests} The authors declare that they have no competing interests.

\bibliographystyle{splncs04}
\bibliography{Paper-1334}

\end{document}

\title{Supplementary Materials: \\ Diff-VPS: Video Polyp Segmentation via a Multi-task Diffusion Network with Adversarial Temporal Reasoning}
\author{Yingling Lu, Yijun Yang, Zhaohu Xing, Qiong Wang, Lei Zhu}

\institute{}
\maketitle              % typeset the header of the contribution

\renewcommand{\floatpagefraction}{.9}
\begin{figure}[!htbp]
    \centering
    \includegraphics[width=\textwidth]{image/test.jpg}
    \caption{\textbf{Overview of Diff-VPS inference phase.} Given a test clip of $\delta+1$ frames, the model starts with a random noise map sampled from a Gaussian distribution and gradually refines the prediction. To speed up the inference, we adopt the paradigm of DDIM. In each sampling step $t$, the random noise $M_i^{T}$ or the predicted noisy map $\hat{M}_i^{T-\Delta}$ from the last step is fused with the conditional feature map and sent to the frozen denoising head for mask prediction. }
    \label{fig:chart}
\end{figure}

\begin{figure}[!htbp]
    \centering
    \includegraphics[width=\textwidth]{image/Appendix Qualitative.jpg}
    \caption{Qualitative results on SUN-SEG hard-seen dataset case 83 .}
    \label{fig:new_chart}
\end{figure}